%% file: main.tex
\documentclass[conference]{IEEEtran}
\usepackage{times}
\pdfoutput=1

\usepackage[numbers]{natbib}
\usepackage{multicol}
\usepackage[bookmarks=true]{hyperref}
\usepackage{graphicx}
\usepackage{amsmath}
\usepackage{amsfonts}
\usepackage{pythonhighlight}

\begin{document}

\title{Enhancing Embodied Object Detection through Language-Image Pre-training and Implicit Object Memory}

\author{Nicolas Harvey Chapman, Feras Dayoub, Will Browne, Christopher Lehnert}



%

\maketitle
\begin{abstract}
Deep-learning and large scale language-image training have produced image object detectors that generalise well to diverse environments and semantic classes. However, single-image object detectors trained on internet data are not optimally tailored for the embodied conditions inherent in robotics. Instead, robots must detect objects from complex multi-modal data streams involving depth, localisation and temporal correlation, a task termed \textit{embodied object detection}. Paradigms such as Video Object Detection (VOD) and Semantic Mapping have been proposed to leverage such embodied data streams, but existing work fails to enhance performance using language-image training. In response, we investigate how an image object detector pre-trained using language-image data can be extended to perform embodied object detection. We propose a novel implicit object memory that uses projective geometry to aggregate the features of detected objects across long temporal horizons. The spatial and temporal information accumulated in memory is then used to enhance the image features of the base detector. When tested on embodied data streams sampled from diverse indoor scenes, our approach improves the base object detector by 3.09 mAP, outperforming alternative external memories designed for VOD and Semantic Mapping. Our method also shows a significant improvement of 16.90 mAP relative to baselines that perform embodied object detection without first training on language-image data, and is robust to sensor noise and domain shift experienced in real-world deployment. 
\end{abstract}

\IEEEpeerreviewmaketitle

\input{Sections/1.0_Introduction}
\input{Sections/2.0_RelatedWork}

\input{Sections/3.0_Preliminaries}
\input{Sections/4.0_Method}
\input{Sections/5.0_Experiments}
\input{Sections/6.0_Robot_Experiments}

\section{Conclusion} 
\label{sec:conclusion}
In this work, we propose an approach to using language-image pre-training and a novel external memory to enhance embodied object detection performance. The implicit object memory effectively aggregates detection information across long temporal horizons, outperforming alternative external memories designed for VOD and Semantic Mapping. The approach is robust to sensor noise and out-of-distribution scenes, and performs well on a real-world deployment scenario. Lastly, we highlight the detection of dynamic objects and completely unseen classes as promising future research directions.



\bibliographystyle{plainnat}
\bibliography{main}

\newpage
\input{Sections/7.0_Appendix}

\end{document}

%% file: Sections/1.0_Introduction.tex
\section{Introduction}

Object detection systems can be used by robots to obtain a semantic understanding of the environment. Developments in deep-learning and large scale language-image training have produced image object detectors that generalise well to diverse environments and semantic classes \cite{vl_distil, exploiting_unlabeled, regionclip, detic, open_vocab_detr, segment_anything}. However, object detectors trained on internet data are not optimally tailored for the embodied conditions inherent in robotics \cite{seal, eal_semseg, self_improving, move_to_see, interactron}. Robots generate long data streams of spatially and temporally correlated data, commonly containing depth and localisation information in addition to images \cite{voxblox, vlmap}. Optimally performing image object detection using this embodied data stream, a task referred to as \textit{embodied object detection}, is the focus of this work.


Several paradigms exist for using an embodied stream of data to detect and localise objects. In Video Object Detection (VOD), the dominant approach is to aggregate image features across sequential frames to deal with challenges such as occlusion, motion-blur and object appearance change \cite{box_1, feature_flow_1, feature_att_1, ogem, mamba, transformer_mem}. However, these methods are not designed to leverage modern object detection systems trained using language-image data. Further, VOD focusses on videos with limited viewpoint shift and relatively few objects per image \cite{traffic_dataset, imagenetvid}, while depth and localisation information are ignored. In contrast, Semantic Mapping and 3D Object Detection make use of embodied data streams to detect and localise objects in 3D space \cite{voxblox, smnet, vlmap, bevformer, bevdet}.  While the resulting geometric representations are advantageous for downstream tasks \cite{ego_map, smnet, vlmap, object_goal}, they are rarely used to improve image object detection \cite{ctmap, seal}. Work is thus needed to understand how an embodied data stream should be used to improve the powerful representations learnt by modern image-based object detectors.

\begin{figure}[!t]
    \centering
    \includegraphics[width=1\columnwidth]{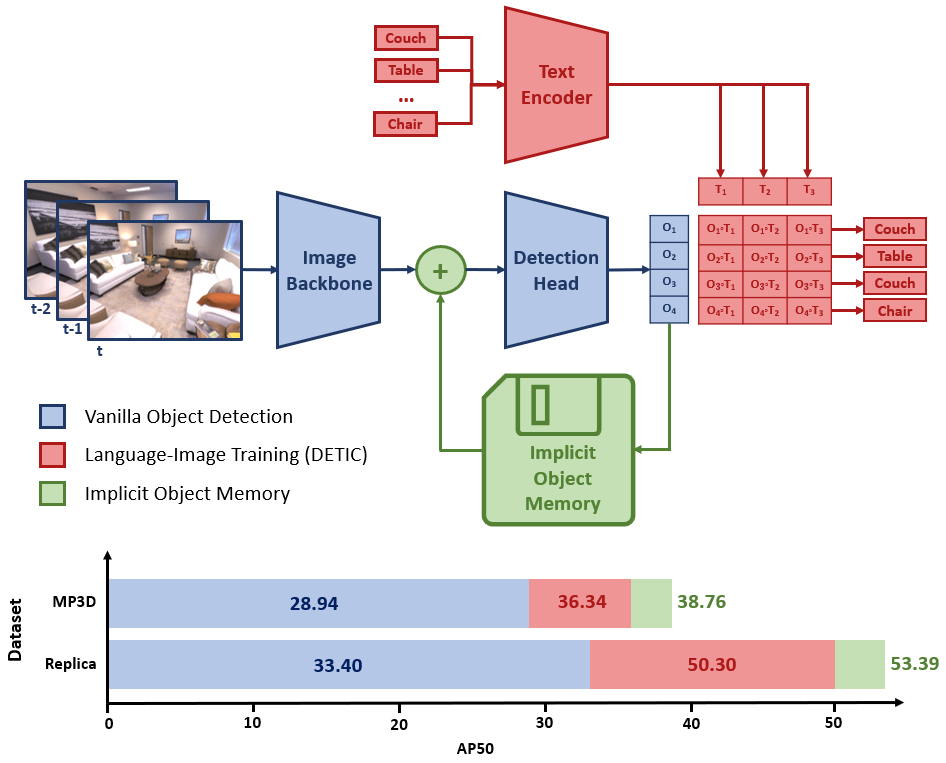}
    \vspace{-2em}
    \caption{Our proposed method for enhancing embodied object detection with language-image training and a novel external memory (top). Our implicit object memory uses projective geometry to aggregate the features of detected objects across long temporal horizons. The spatial and temporal information stored in this external memory is then used to enhance the image features of the base detector. We evaluate our method on embodied data streams sampled from two datasets of indoor scenes (bottom). Relative to performing vanilla object detection (blue), the inclusion of language-image pre-training (red) leads to an increase of 7.40 mAP on Matterport3D and 16.90 mAP on the Replica test sets. Adding our implicit object memory (green) results in a further 2.46 mAP and 3.09 mAP improvement respectively.
    }
    \vspace{-1.5em}
    \label{fig:HookFigure}
\end{figure}


In response, we investigate how an image object detector trained using language-image data \cite{detic} can be extended to perform embodied object detection. We propose the use of an external memory to enhance the image features of the base detector using spatial and temporal information (Figure \ref{fig:HookFigure}). This approach has been used in VOD to separate the storage of past observations from the representation space of the model, avoiding redundant computation and yielding state-of-the-art performance \cite{ogem, mamba, transformer_mem}. However, we find existing approaches aimed at aggregating image features across time fail to leverage long-term information.

Motivated by this, we propose a method that uses projective geometry to maintain a novel implicit object memory. By aggregating implicit object features across spatial dimensions as in Semantic Mapping \cite{vlmap, smnet}, long-term observation of the environment can be utilised to improve object detection performance. This external memory is then used to enhance the image features of an object detector trained using internet scale language and image data \cite{detic}. Our method leads to improved object detection when tested on embodied data streams sampled from indoor scenes \cite{replica, mp3d} (Figure \ref{fig:HookFigure}), outperforming alternative external memories designed for VOD and Semantic Mapping. Importantly, the approach is robust to the sensor noise and data shift commonly experienced during robotic deployment \cite{udaod1, udaod2}. We also demonstrate significant improvement relative to baselines that perform embodied object detection without first training on language-image data (Figure \ref{fig:HookFigure}). 


To summarise, this paper makes the following contributions:
\begin{itemize}
    \item A method for using external memory to enhance the feature space of an object detector trained on internet scale language-image data.
    \item A method for using projective geometry to maintain an implicit external memory that captures long-term object dependencies.
    \item A detailed evaluation of our approach on the task of embodied object detection, where data streams are collected by a robot in indoor scenes.
\end{itemize}

%% file: Sections/2.0_RelatedWork.tex
\section{Related Work}
\subsection{Embodied Object Detection}
The task of \textit{embodied object detection} involves performing image object detection on the embodied data stream produced by a robot as it moves through an indoor environment. Along with images, we assume the robot has access to localisation and depth information. Existing work in this space is motivated by embodied exploration and navigation tasks \cite{seal, embodied_exploration}, which predominately rely on pre-trained image object detectors. However, as pointed out in the Embodied Active Learning (EAL) literature \cite{seal, eal_semseg, self_improving, move_to_see, interactron}, object detectors trained using internet data are not optimally tailored for embodied conditions. In response, EAL methods actively control the path of the robot to produce an unsupervised learning signal for updating the weights of the model. However, despite using depth and localisation information to fine-tune the model, these approaches rely solely on images to perform detection. Furthermore, existing work in EAL fails to leverage models trained using language-image data.

Our work instead focusses on how to produce an improved embodied object detector to replace the image-based models used in EAL and embodied navigation. Specifically, we aim to incorporate spatial and temporal information into an object detector pre-trained using large scale language-image data. Our proposed solution combines three core paradigms that are currently studied separately in the literature. Firstly, we use an open-vocabulary object detector trained on language-image data as the base model. Secondly, we investigate how spatial-temporal information can be incorporated into the base detector via an external memory, as done in VOD. Lastly, we consider methods that use projective geometry to build semantic representations of a scene, such as Semantic Mapping and 3D Object Detection. 

\subsection{Language-Image Pre-training for Object Detection}
The availability of captioned images has allowed the joint training of image and text encoders using natural language supervision. The seminal work in this space is Contrastive Language-Image Pretraining (CLIP), which uses a contrastive loss to produce similar embeddings for image-text pairs \cite{clip}. By quantifying the similarity of images and text descriptions, the resulting encoders can be used to perform image classification with open-vocabulary classes \cite{clip}. As natural language descriptions are highly generalisable, CLIP also exhibits superior resistance to domain shift compared with supervised pre-training methods \cite{clip, grounding_with_text}. These benefits have been extended to object detection by training models to produce region features in the CLIP embedding space \cite{vl_distil, exploiting_unlabeled, regionclip, detic, open_vocab_detr, segment_anything}. In particular, DETIC scales this approach using a combination of object detection and image classification datasets to train on 21k classes \cite{detic}. A shared image-text representation space has also been used to train multimodal Large Language Models (LLMs), which perform general sequence prediction tasks given language and image inputs \cite{mdetr, contextual_detection, palme}. While these methods combine traditional object detection with complex reasoning capacity, they inherit the high computational complexity of LLMs \cite{palme}. For this reason, we focus on methods that train light-weight object detection architectures using large scale language-image data \cite{detic}. We explore how these systems can be improved for use in robotic deployment environments, with a particular focus on how spatial and temporal information can be used to enhance the image representation space.

\subsection{Video Object Detection}
Video Object Detection (VOD) aims to leverage temporal information from videos to overcome deteriorated image quality due to occlusions, motion-blur and object appearance change. Box-level approaches to VOD aim to associate detected boxes through time to improve detection \cite{box_1, box_2, box_3, box_4}. However, such methods rely on objects being detected across many frames to perform association, and cannot be trained end-to-end. Feature based methods instead aim to enhance the image features of the current frame using those from previous frames. This is done using optical flow to align past images with the current frame \cite{feature_flow_1, feature_flow_2, feature_flow_3, feature_flow_4, feature_flow_5}, attention to fuse similar features \cite{attention, feature_att_1, feature_att_2, feature_att_3, feature_att_4} or modelling object relations across frames to perform instance-level enhancement \cite{SELSA, mega, relation_networks}.  Due to view-point shift and appearance change across long time horizons, these methods generally struggle to model long-term spatial-temporal dependencies. 

In response, methods have been proposed to enhance image features by using attention-based heuristics to read and write to an external memory \cite{ogem, mamba, transformer_mem}. The use of an external memory avoids much redundant computation by separating the storage of past observations from the representation space of the model \cite{ogem}. Such methods are better at modelling long-term spatial-temporal dependencies, leading to state-of-the-art results on VOD benchmarks. External memory has the further advantage of being agnostic to the object detection architecture, meaning that it can benefit models trained using internet scale image and language data. However, existing approaches to video object detection have not been applied to such models. Furthermore, VOD is generally tested on datasets with limited viewpoint shift \cite{traffic_dataset} and few objects per image \cite{imagenetvid}, and fails to utilise depth and localisation information. Consequently, we find that in robotic deployment environments, existing external memories aimed at aggregating image features through time struggle to utilise long-term information.


\subsection{Semantic Mapping and 3D Object Detection}
Embodied object detection is also related to paradigms that leverage pose and depth information to detect objects in 3D space. In Semantic Mapping, object detections and projective geometry are used to assign semantic labels to geometric representations \cite{voxblox, vlmap}. Recent methods combine the benefits of neural representation learning and projective geometry to learn an implicit representation of a scene \cite{smnet, trans4map, ego_map}. These implicit representations provide a useful spatial memory for planning and navigation tasks \cite{object_goal, ego_map, smnet, vlmap}, but have not been used to improve image object detection. Projective geometry is also used in 3D Object Detection, where image features are fused from multiple views to detect objects in 3D space \cite{3d_object_detection}. Recent approaches build implicit Bird's Eye View (BEV) features, which can be decoded into 3D bounding boxes, semantic classifications and a semantic map \cite{bevformer, bevdet, recurrent_fusion}. However, these implicit scene representations are not used to improve image object detection systems or enhance the representations learnt during language-image training. 



%% file: Sections/3.0_Preliminaries.tex
\section{Preliminaries}
\subsection{Problem Formulation}
Embodied object detection considers the scenario where a robot collects from the environment a continuous sequence of RGB-D images and robot pose information. At each time $t$, a new image $I_{t}$ and robot pose $Q_{t}=\{x_{t},y_{t},z_{t},\theta_{t}\}$ are sampled. Given this data as input, the goal is to sequentially predict the bounding-box locations $b^{t}\in\mathbb{R}^{k \times 4}$ and class-specific likelihood scores $s^{t}\in\mathbb{R}^{k \times C}$ of objects $k$ from $C$ classes in each image $I_{t}$.


\subsection{Object Detection with Language-Image Embeddings}
The base object detection models used in this work leverage a shared language-image embedding space to generate class-specific likelihood scores. The first component of such detectors is a feature extraction backbone $f$ that converts images $I_{t}$ to pixel feature maps $z_{p}^{t}\in\mathbb{R}^{w \times h \times d_{1}}$. These features maps are then processed by a detection head $g$ to produce a set of corresponding bounding-box locations $b^{t}\in\mathbb{R}^{k \times 4}$, object masks $m^{t}\in\mathbb{R}^{k \times w \times h \times C}$, objectness scores $o^{t}\in\mathbb{R}^{k}$ and object features $z_{o}^{t}\in\mathbb{R}^{k \times d_{2}}$, where $k$ defines the number of object proposals per image, $w$ and $h$ refer to the height and width of the feature map and $d_{1}$ and $d_{2}$ refer to the size of different feature vectors. The output of the object detector given an image $I_{t}$ is thus defined by the following equation:

\begin{equation}
    \{b^{t},m^{t},o^{t},z_{o}^{t}\} = g \circ f\left(I_{t}\right)
\end{equation}

The language-image embedding space \cite{clip, detic} is then used to convert the object features $z_{o}^{t}$ to class-specific classification scores $s_{l}$. Firstly, the class labels $L$ are processed by the pre-trained CLIP text encoder $T$. The raw output of the text encoder is projected into a lower dimensional space and normalised to produce the final class-specific embeddings $z_{l} \in\mathbb{R}^{C \times d_{2}}$:

\begin{equation}
z_{l} = \left\| T(L) \cdot W_{t} \right\|_{2}
\end{equation}

where $W_{t}$ is a learnt projection matrix. Next, the cosine similarity between the object features $z_{o}^{t}$ and the class-specific embeddings $z_{l}$ is calculated and converted to a likelihood score by applying the sigmoid function $\sigma$. The final class-specific likelihood score $s^{t}$ is the geometric mean of this likelihood score and the objectness score returned by the detection head: 

\begin{equation}
s^{t} = \sqrt{\sigma\left(z^{t}_{o} \cdot z_{l}\right)*o^{t}}
\end{equation}

%% file: Sections/4.0_Method.tex
\section{Method}
\subsection{Spatial Structure of Implicit Object Memory}
The implicit object memory $M\in\mathbb{R}^{a \times l \times d_{2}}$ is structured as a variable-sized 2-dimensional tensor, where each element represents a 0.2m $\times$ 0.2m square on the ground-plane of a scene. The size of $M$ is dependent on the breadth $a$ and length $l$ of the scene, as well as the size of the feature $d_{2}$ stored at each location. The memory feature at time $t$ and element $(u, v)$ is thus referred to as $M^{t}_{u,v}$. In addition, a second matrix $V\in\mathbb{R}^{a \times l}$ is used to store the number of times that each location in memory is viewed by the robot.

\begin{figure*}[!t]
    \centering
    \includegraphics[width=1.99\columnwidth]{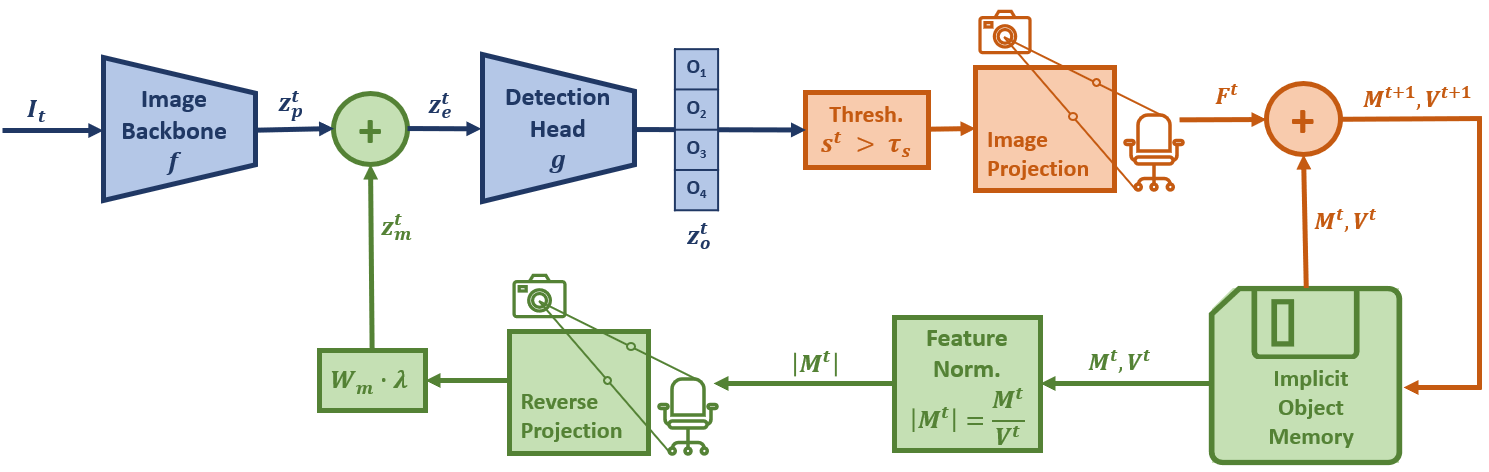}
    \vspace{-0.5em}
    \caption{Our implicit object memory for enhancing the feature space of a base object detector with spatial and temporal information. Our memory write operation (orange) first selects all objects predicted by the base detector with class-specific likelihood score $s^{t}$ greater than some threshold $\tau_{s}$. Projective geometry is then used to map the object features $z_{o}^{t}$ from their location in the image-frame to the corresponding location in spatial memory. The projected object features $F^{t}$ are then summed at each location with those already in implicit object memory $M^{t}$ to generate the updated memory matrix $M^{t+1}$. A count of the number of times each memory location is viewed by the robot $V^{t}$ is also incremented. The read operation involves normalising each feature in implicit object memory $M^{t}$ based on how many times the location has been viewed. The normalised features $\left|M^{t}\right|$ are projected back into the image-frame using the reverse image projection. The resulting egocentric memory features $z_{m}^{t}$ are passed through a linear projection layer $W_{m}$ to align them with the original pixel features $z^{t}_{p}$. To produce the final enhance pixel features $z^{t}_{e}$ the egocentric memory features are weighted by a coefficient $\lambda$ and summed with the original pixel features.}
    \vspace{-1.5em}
    \label{fig:method}
\end{figure*}

\subsection{Writing to Implicit Object Memory}

The proposed implicit object memory is updated given the output of the object detector using the following write operation, shown in more detail in Figure \ref{fig:method}:

\begin{equation}
    \{M^{t+1}, V^{t+1}\} = write\left(M^{t}, V^{t}, z^{t}_{o}, o^{t}, m^{t}\right)
\end{equation}

The first step of this operation is to select all objects with a class-specific likelihood score above some threshold $\tau_{s}$. The class-specific likelihood scores $s^{t}$ are calculated via Eq. 3 and only confident object features $z^{t}_{c}$ that satisfy $s^{t}>\tau_{s}$ are used to update memory. 

Projective geometry is then used to map the location of confident objects in the image-frame to the corresponding location in spatial memory. The masks $m^{t}$ output by the detector are firstly used to associate confident object features $z^{t}_{c}$ to pixels $(i,j)$. For each image pixel associated with an object, a ray is shot from the camera center through the pixel to a depth of $d_{i,j}$ to calculate its 3D position relative to the camera. This 3D position is then converted to a global position using the camera extrinsic matrix $R$, which can be calculated knowing the current pose of the robot. Approximating the camera as a pinhole with intrinsic matrix $K$, this projection process can be written as:

\begin{equation}
\begin{bmatrix} x \\ y \\ z \end{bmatrix} = d_{i,j}R^{-1}K^{-1}
\begin{bmatrix} i \\ j \\ 1 \end{bmatrix}
\end{equation}

The global coordinates $(x,y)$ are then used to lookup the associated location in memory $(u,v)$. Features that are projected to the same location in memory are averaged, and locations not associated with detected objects are filled with zeros. The result is a matrix of projected object features $F^{t}\in\mathbb{R}^{a \times l \times d_{2}}$ with the same shape as the memory matrices $M^{t}$ and $V^{t}$. The projected features at each location are then summed with those in $M^{t}$ to generate the updated memory matrix:

\begin{equation}
    M^{t+1}_{u,v} = M^{t}_{u,v} + F^{t}_{u,v}
\end{equation}

Lastly, the value of $V^{t}$ at any location $(u,v)$ that is visible in the current image is incremented:

\begin{equation}
    V^{t+1}_{u,v} = V^{t}_{u,v} + 1
\end{equation}

\subsection{Pixel Feature Enhancement with Implicit Object Memory}
A novel memory read operation is proposed to enhance the pixel features $z^{t}_{p}$ of the base object detector with information stored in the implicit object memory:

\begin{equation}
    z^{t}_{e} = read\left(z^{t}_{p}, M^{t}, V^{t}\right)
\end{equation}

The complete read operation is summarised in Figure \ref{fig:method}. Firstly, we normalise each memory feature $M^{t}_{u,v}$ based on how many times the location $(u,v)$ has been viewed. The goal of this operation is to generate features with large magnitude in locations where objects are regularly observed. In contrast, locations that are usually ignored as background will have low magnitude. To generate the normalised implicit object feature $\left|M^{t}_{u,v}\right|$, we scale the original feature $M^{t}_{u,v}$ by the inverse of the view-count $V^{t}_{u,v}$:

\begin{equation}
    \left|M^{t}_{u,v}\right| = \frac{M^{t}_{u,v}}{V^{t}_{u,v}}
\end{equation}

The normalised features are then projected back into the image frame using the inverse of Eq. 5 to generate egocentric memory features $z^{t}_{m}\in\mathbb{R}^{w \times h \times d_{2}}$. The pixel features and egocentric memory features are now aligned along the spatial dimension, but have different feature sizes $d_{1}$ and $d_{2}$. To also align the feature dimension, the egocentric memory features are passed through a linear projection layer $W_{m}\in\mathbb{R}^{d_{2} \times d_{1}}$. To generate the final enhanced features $z^{t}_{e}$, the weighted sum of the egocentric memory features $z^{t}_{m}$ and pixel features $z^{t}_{p}$ is calculated, with the coefficient $\lambda$ used to control the relative contribution of the memory features:

\begin{equation}
    z^{t}_{e} = z^{t}_{m} \cdot W_{m} \cdot \lambda + z^{t}_{p}
\end{equation}

As shown in Figure \ref{fig:HookFigure}, the enhanced pixel features can then be passed to the detection head and the network trained using a standard object detection loss. The only learnable parameters added to the model in this process are those in the linear projection matrix $W_{m}$. 

\subsection{Baseline External Memories}
\textbf{Explicit Object Memory.}
Motivated by existing work in Semantic Mapping \cite{object_goal, ego_map, smnet, vlmap, voxblox, trans4map}, we implement an embodied object detection baseline that uses an explicit semantic map as external memory. To generate the semantic map $S\in\mathbb{R}^{a \times l}$, we convert the feature at each location in our implicit object memory to an explicit classification. 

Firstly, a binary occupancy map $O\in\mathbb{R}^{a \times l}$ is produced by finding locations where an object has been regularly detected. We use the magnitude of each memory feature $M^{t}_{u,v}$ as a proxy for the number of times an object has been detected at location $(u,v)$. We divide this value by corresponding view-count $V^{t}_{u,v}$ to account for the degree of observation of each location, giving the following occupancy ratio:

\begin{equation}
    r_{o} = \frac{\left|M^{t}_{u,v}\right|}{V^{t}_{u,v}}
\end{equation}

If this ratio is above a given threshold $\tau_{o}$, the location is considered occupied by an object:

\begin{equation}
O^{t}_{u,v} = 
\begin{cases}
  0 & \text{if } r_{o} < \tau_{o}, \\
  1 & \text{if } r_{o} \geq \tau_{o}
\end{cases}
\end{equation}



For each location considered to contain an object, we can  leverage the shared language-image embedding space to perform classification. The cosine similarity between the normalise object feature $\left|M^{t}_{u,v}\right|$ and the class-specific embeddings $z_{l}$ is calculated, and the object with the highest similarity is then stored in the explicit semantic map:

\begin{equation}
S^{t}_{u,v} = 
\begin{cases}
  0 & \text{if } O^{t}_{u,v} = 0, \\
  max\left(\left|M^{t}_{u,v}\right| \cdot z_{l} \right) & \text{if } O^{t}_{u,v} = 1
\end{cases}
\end{equation}

To use the explicit semantic map for embodied object detection, we simply replace the explicit classification $S^{t}_{u,v}$ with the corresponding class-specific embedding from $z_{l}$. The pixel feature enhancement operation can then be applied with the explicit object features as per Eq. 8.

\textbf{Implicit Pixel Memory.}
As opposed to using object features, various implicit semantic mapping approaches \cite{smnet, trans4map, ego_map} aggregate pixel features through time to build an implicit representation of a scene. Motivated by these methods, we propose an approach for using an implicit pixel memory $P\in\mathbb{R}^{a \times l \times d_{3}}$ for embodied object detection. 

As in \citet{smnet}, pixel features $z^{t}_{p}$ extracted by the image backbone are projected into the spatial memory using Eq. 5. The resulting projected pixel features $G^{t}\in\mathbb{R}^{a \times l \times d_{1}}$ are merged with the implicit pixel memory from the previous time-step using a small recurrent network:

\begin{equation} \label{eq10}
P^{t}_{u,v} = RNN(G^{t}_{u,v}, P^{t-1}_{u,v})
\end{equation}

As in \citet{smnet}, the weights of the RNN are shared across all locations in memory. The pixel feature enhancement operation can then be applied with the implicit pixel features as per Eq. 8.

\textbf{MAMBA External Memory.}
We additionally add the external memory from MAMBA \cite{mamba} to the base detector. This method is representative of the attention-based external memories that are state-of-the-art on VOD benchmarks \cite{ogem, mamba, transformer_mem}. MAMBA maintains two distinct memory banks that store previously observed pixel and instance features respectively. Given a new image, a subset of the memory features are randomly sampled to perform feature enhancement. A generalised enhancement operation transforms the current image features by recursively attending to the sampled memory features. The pixel and instance features associated with confident detections are then added to the memory banks. We use this approach to enhance the pixel features $z^{t}_{p}$ and the object features $z^{t}_{o}$ of the base detector.

%% file: Sections/5.0_Experiments.tex
\section{Dataset Experiments}
\subsection{Experimental Settings}
\textbf{Datasets.}
We use the Habitat simulator \cite{habitat} with Matterport3D scans \cite{mp3d} to generate data for training and evaluating the embodied object detection task. This setup allows for the rendering of semantic labels and RGB-D images from any viewpoint in realistic indoor environments. Consequently, this setup is common in recent semantic mapping \cite{smnet, vlmap} and embodied active learning \cite{seal, eal_semseg, self_improving, move_to_see} works. For additional testing data, we use renderings of indoor scenes from the Replica dataset \cite{replica}. 

The Matterport3D dataset contains scans for 90 indoor environments, with dense annotations of 40 object classes. For our experiments, we use 15 commonly occurring object categories of \textit{bed, towel, fireplace, picture, cabinet, toilet, curtain, table, sofa, cushion, bathtub, chair, chest of drawers} and \textit{tv monitor}. Utilising the same dataset split as \citet{smnet}, we use 61 scenes for training, 7 for validation and 22 for testing. The Replica dataset contains 19 indoor scenes with semantic annotations for 88 classes. We generate a class mapping from the Replica to Matterport3D classes (see supplementary material) and use all scenes in the Replica dataset as an additional test set. 

\textbf{Preparation for Embodied Object Detection.}
To generate data for training and evaluating embodied object detection, we randomly generate short paths in each scene, termed episodes. At each step in an episode, RGB-D images, robot pose and object bounding boxes are extracted using the Habitat simulator. As in \citet{smnet}, we choose an episode length of 20 steps to produce a diverse training and testing set while keeping computational costs low. At inference time, episodes from the same scene can be sequentially processed to assess long-term deployment in an environment. We collect 50 episodes for each scene, resulting in 3050 training, 350 validation, 1100 testing and 950 replica episodes.

\textbf{Base Object Detection Architecture.}
For all experiments, we use the Feature Pyramid Network (FPN) \cite{fpn} version of the CenterNet2 detector \cite{centernet2} with a ResNet50 backbone \cite{resnet}. This architecture was used to train DETIC models for real-time performance \cite{detic}, and is thus preferred over transformer based architectures \cite{swin} for robotics applications. This architecture uses a pixel feature dimension of 256, and an object feature dimension of 512.

\textbf{Baselines.} 
We compare our implicit object memory to the baseline external memories introduced in Section IV.D. 
We also compare various approaches to pre-training the base object detector:
\begin{itemize}
    \item \textbf{Pre-trained DETIC.} Large scale language-image training as per \citet{detic}. We simply use the weights generated by the authors when training the base detector on LVIS \cite{lvis}, COCO \cite{coco} and ImageNet-21k \cite{imagenet21k}.
    \item \textbf{Fine-tuned DETIC.} The network is initialised with the pre-trained DETIC weights, and fine-tuned on the Matterport3D train set.
    \item \textbf{Vanilla Training.} The backbone is initialised from ImageNet-21k weights \cite{imagenet21k} and the detection head intialised randomly. The model is then trained on the Matterport3D train set.
\end{itemize}

\textbf{Implementation Details.}
We use the implementation provided by DETIC to train all models \cite{detic}, which leverages Detectron2 \cite{detectron2}. A batch size of 2 episodes (40 samples) is used to train all methods, as this is the maximum batch size that fits on a single A100 GPU. SGD is used as the optimiser for all experiments. When performing vanilla training, a learning rate of 0.0001 is used. When fine-tuning from DETIC weights, we use a learning rate of 0.00001. The base model is then fine-tuned end-to-end with the external memory, using a learning rate of 0.00001 for layers in the base detection network and 0.0001 for new layers associated with the external memory. All models are trained for a total of 10000 iterations on the Matterport3D train set, with a checkpoint saved every 1000 iterations. To select a model for testing, the checkpoint with the highest performance on the validation set is selected. Where an earlier checkpoint reaches within 0.3 mAP of the best performing model, it is instead used to mitigate overfitting. Additionally, the following settings are used to implement the external memories:
\begin{itemize}
    \item \textbf{Implicit Object Memory.} We pre-compute the implicit object memory using the base object detector to speed up training. A confidence threshold of 0.3 is used to select detections for updating memory. The weighting coefficient for the implicit object memory is set to 5 to optimise performance on the test set (Figure \ref{fig:hyperparams}).
    \item \textbf{Explicit Object Memory.} We also pre-compute the explicit object memory for training. A confidence threshold of 0.3 is also used to select confident defections for updating memory, and the threshold applied to create the occupancy map is 0.4. The weighting coefficient for the explicit object memory is set to 100 to optimise performance on the test set (Figure \ref{fig:hyperparams}).
    \item \textbf{Implicit Pixel Memory.} We closely follow the publicly available implementation of \citet{smnet} to build the implicit pixel memory, using a single layer GRU with hidden dimension 256 to update the spatial memory. Because this GRU must be trained end-to-end, the implicit pixel memory cannot be pre-computed. The weighting coefficient used to combine implicit pixel memory and image features is set to 20 to optimise performance on the test set (Figure \ref{fig:hyperparams}).
    \item \textbf{MAMBA External Memory.} We closely follow the publicly available implementation of \citet{mamba} to implement the MAMBA external memory, changing only the optimiser, batch size and confidence threshold to align with the other baselines.
\end{itemize}

\textbf{Evaluation.}
Standard mean average precision at an Intersection over Union (IoU) of 50 (AP50) is used to perform evaluation on the Matterport3D and Replica test sets. By default, the 50 episodes from each test scene are processed sequentially to simulate long-term deployment in an environment, with the external memory allowed to persist across episodes. 

\subsection{Comparison of Object Detection Pre-training}
We firstly evaluate alternative approaches to pre-training the base object detector for embodied object detection (Table \ref{table1}). The results highlight that language-image training alone is not sufficient to produce an optimal model. Fine-tuning DETIC on the training set generates an increase in performance on both the Matterport3D (+18.76 mAP) and Replica (+13.10 mAP) test sets (Table \ref{table1}). However, language-image pre-training remains valuable, with the fine-tuned model performing notably better than vanilla training (Table \ref{table1}). This is particularly true on the Replica dataset, where the fine-tuned DETIC model is stronger by 16.90 mAP and the pre-trained DETIC model by 3.80 mAP. Evidently, performing language-image pre-training leads to a detector that is more robust to domain shift between the training and testing datasets \cite{grounding_with_text, clip}. This is particularly important in robotics applications, where labelled data is difficult to acquire and deployment environments can differ significantly \cite{udaod1, udaod2}.

\begin{figure}[!t]
    \centering
    \includegraphics[width=0.94\columnwidth]{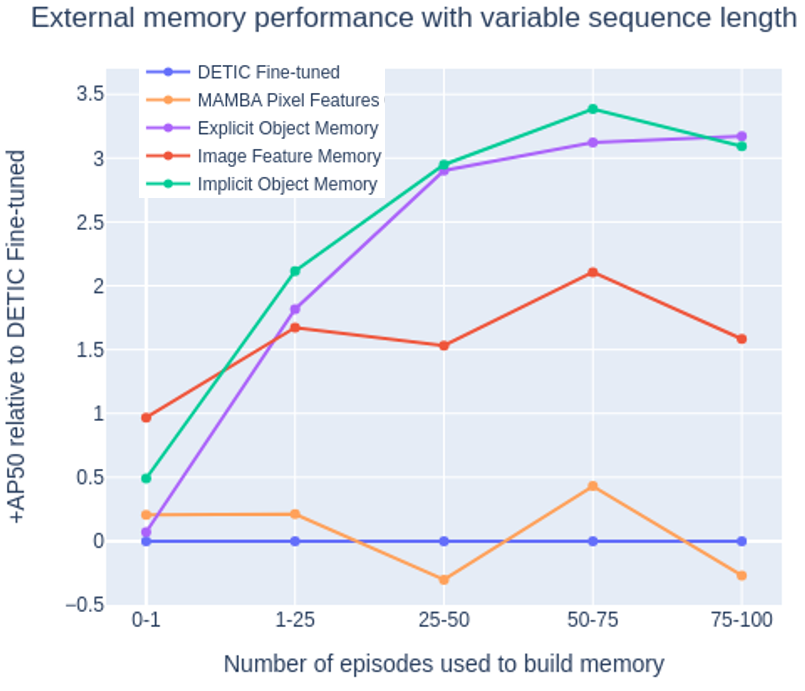}
    \vspace{-1em}
    \caption{Performance of alternative external memories across different sequence lengths on the Matterport3D test set. All external memories are training with fine-tuned DETIC as the base model. To generate results across 100 episodes, the 50 episodes from each scene are processed twice in sequential order, with the external memory allowed to persist across all 100 episodes. Performance is calculated separately for different stages to investigate if the external memory becomes more useful as the number of episodes increases. Performance across a single episode is also reported, whereby the memory is instead reset at the end of every episode.
    }
    \label{fig:longterm}
\end{figure}

\subsection{Comparison of Alternative External Memories}
Given the evident value of large scale language-image pre-training for embodied object detection, our next set of experiments explore how long-term spatial-temporal information can be incorporated into such models. We assess the ability of alternative external memories to aggregate information across time and improve embodied object detection performance. The implicit object memory outperforms other methods designed for Semantic Mapping and VOD, particularly on the Replica dataset where there is a 2.94 mAP improvement relative to the next best model (Table \ref{table2}). Furthermore, our method effectively leverages long-term observation of the environment, performing optimally when greater than 50 episodes are used to build external memory (Figure \ref{fig:longterm}).


The explicit object memory also performs well on the Matterport3D test set (Table \ref{table2}) and can similarly be used to capture long-term information (Figure \ref{fig:longterm}). The key difference between the two approaches is the conversion of the implicit object feature at each location to an explicit classification. Despite this relatively small change, the performance of the explicit object memory degrades significantly on the Replica test set (Table \ref{table2}). Evidently, an implicit memory is important for strong out-of-distribution performance and is thus preferred to a generic semantic mapping approach. We attribute this to the ability of our implicit object features to express uncertainty in object classification by aggregating the embeddings of different classes across time, and the robustness of the object features learnt during language-image pre-training.

\begin{table}[t]
\centering
\caption{Comparison of alternative approaches to pre-training the base object detector on the Matterport3D and Replica test sets.}
\begin{tabular}{l|l|l}
\hline
                                                 & MP3D              & Replica      \\ \hline
Fine-tuned DETIC + Implicit Object Memory        & \textbf{38.76}             & \textbf{53.39}        \\ \hline
Fine-tuned DETIC                                 & 36.34             & 50.30        \\ \hline
Vanilla Training + Implicit Object Memory        & 30.17             & 34.48        \\ \hline
Vanilla Training                                 & 28.94             & 33.40        \\ \hline
Pre-trained DETIC                                & 17.58             & 37.20        \\ \hline
\end{tabular}
\label{table1}
\vspace{-1em}
\end{table}

The results also emphasise that object-based memory is more effective than implicit semantic mapping approaches that aggregate image features through time. The addition of implicit pixel memory to the fine-tuned DETIC model results in only a marginal improvement on the test sets (Table \ref{table2}). Further, while the implicit pixel memory performs well when memory is reset after every episode, it does not benefit significantly from continued observation of a scene (Figure \ref{fig:longterm}). We thus find that pixel-based memories may be useful for modelling dependencies between sequential images, as opposed to long-term object relationships. We attribute this to the high computational cost of manipulating low-level pixel features, and the variance of object appearance across vastly different viewpoints. In contrast, object features are more abstracted and are viewpoint invariant.

Lastly, we find that the attention-based external memory of MAMBA performs poorly on the embodied object detection task. We attribute this to the greater viewpoint shift and numbers of objects per image in embodied object detection relative to in VOD datasets. Under such conditions, attending across previously detected regions and those in the current image is an overly complex operation. As such, four A100 GPUs are required to train MAMBA with a similar batch size to other methods. Furthermore, the pixel and instance based external memories produce an insignificant improvement when added to the fine-tuned DETIC model (Table \ref{table2}), and there is no evidence of improved performance with larger numbers of episodes (Figure \ref{fig:longterm}). In contrast, our proposed implicit object memory deals with the complexity of embodied object detection by leveraging projective geometry as an inductive bias for enhancing image features.

\begin{figure}[!t]
    \centering
    \includegraphics[width=1\columnwidth]{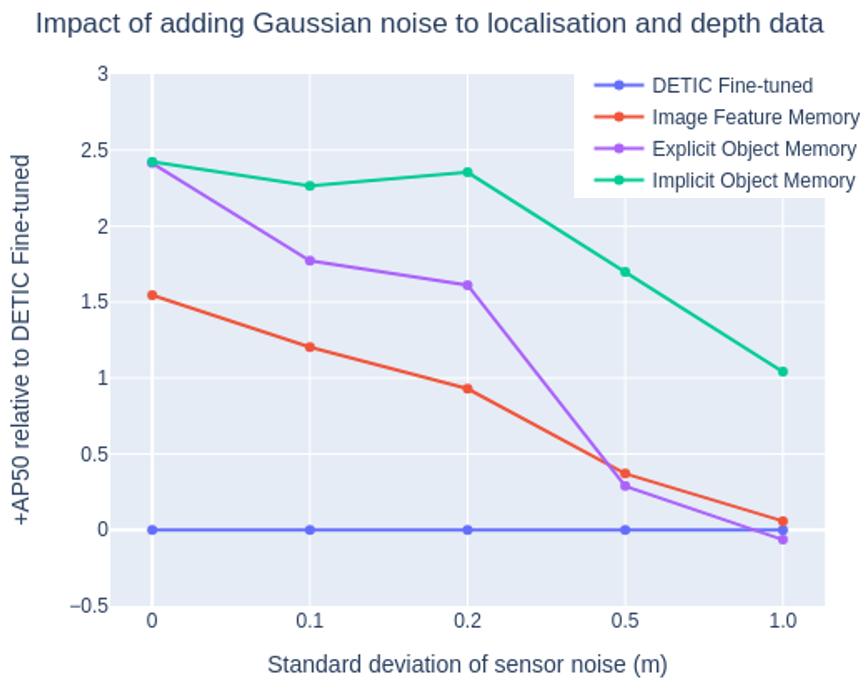}
    \vspace{-2em}
    \caption{Performance of alternative external memories when subjected to noise in depth and localisation. All external memories are training with fine-tuned DETIC as the base model. Gaussian noise with a standard deviation of 0.1m is added to the depth and position reading, and 0.01 radians to the robot heading. The standard deviation is scaled by a factor of 2, 5 and 10 until significant performance degradation is realised.
    }
    \label{fig:noise}
\end{figure}

\begin{figure*}[!t]
    \centering
    \includegraphics[width=1.99\columnwidth]{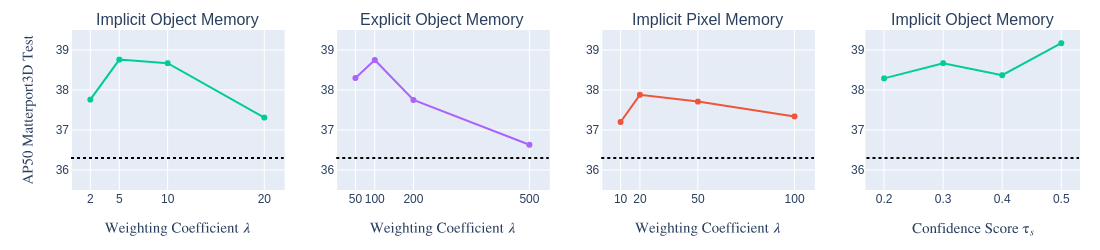}
    \vspace{-1em}
    \caption{Sensitivity of proposed external memories to key hyper-parameters on the Matterport3D test set. In each test, the isolated parameter is swept while keeping all other aspects of implementation constant and fine-tuned DETIC is used as the base model. The dotted line represents the performance of the fine-tuned DETIC model.
    }
    \label{fig:hyperparams}
    \vspace{-1em}
\end{figure*}

\begin{table}[t]
\centering
\caption{Comparison of alternative external memories when added to fine-tuned DETIC. The value in parenthesis refers to the change in performance relative to using fine-tuned DETIC without external memory.}
\begin{tabular}{l|l|l}
\hline
                                            & MP3D                   & Replica                \\ \hline
Implicit Object Memory   & \textbf{38.76 (+2.42)} & \textbf{53.39 (+3.09)} \\ \hline
Explicit Object Memory   & 38.75 (+2.41)          & 49.15 (-1.15)          \\ \hline
Implicit Pixel Memory    & 37.88 (+1.54)          & 50.30 (+0.00)          \\ \hline
MAMBA Pixel and Instance & 36.55 (+0.21)          & 49.40 (-0.90)          \\ \hline
MAMBA Pixel only         & 36.54 (+0.20)          & 50.45 (+0.15)          \\ \hline
Fine-tuned DETIC                            & 36.34                  & 50.30                  \\ \hline
\end{tabular}
\label{table2}
\vspace{-1em}
\end{table}

\subsection{Sensitivity to Depth and Localisation Accuracy}
The Matterport3D and Replica scans used in these experiments provide error free depth and pose information, which is not indicative of real-world sensors. For example, the Intel RealSense D455 camera is a common sensor used in robotics and guarantees an error less than 0.08m at a distance of 4m. Further, state-of-the-art SLAM systems such as ORB-SLAM3 \cite{orbslam3} return localisation accuracy within 0.04m on complex scenes. To be useful for robotics applications, our proposed method should be robust to such levels of noise. We thus add Gaussian noise to the depth, position and heading used to perform geometric projection. As a starting point, we add Gaussian noise with a standard deviation of 0.1m to the depth and position readings, and 0.01 radians to the heading. This noise is scaled until performance degradation is realised. The results show that the proposed implicit object memory is robust to noise in the geometric projection process, suffering only minor performance drop when 0.2m of noise is added to position and depth (Figure \ref{fig:noise}). The explicit object memory is the least robust (Figure \ref{fig:noise}), mirroring its poor performance on out-of-distribution scenes. This further highlights that the ability to encode uncertainty in object classification is important for reliable embodied object detection.

A related limitation of these experiments is the lack
of dynamic objects. The use of projective geometry to read and write to memory assumes that objects will persist in previously observed locations. The implicit object memory is therefore likely to ignore dynamic objects, leaving the embodied object detector to detect them using image features only. While dynamic objects are present in our final experiment with the real robot, a detailed assessment of this problem is left to future work. Methods that predict long-term object dynamics \cite{object_modeling} could also be investigated to enhance our external memory and further improve embodied object detection.

\subsection{Open-vocabulary Performance}
In addition to producing robust and generalisable models, language-image pre-training enables object detection to be performed on novel classes, a task termed open-vocabulary object detection \cite{vl_distil, exploiting_unlabeled, regionclip, detic, open_vocab_detr, segment_anything}. We demonstrate the performance of our proposed implicit object memory on a constrained version of this task, where the class names in the original Replica and Matterport3D datasets are replaced with synonyms at test-time. This  reflects the scenario where a human user defines an object for detection that differs slightly from what the robot has seen during training. The class mapping used in this experiment, which was provided by ChatGPT, can be found in the supplementary material. Our implicit object memory remains the optimal approach under this setting, despite never being trained on the synonymous classes (Table \ref{table3}). Evidently, fine-tuning DETIC on a small subset of classes does not undermine its ability to use open-vocabulary class names. However, there is no evidence that the approach would improve performance on novel classes that differ significantly from those in the training set. Expanding this work to perform embodied object detection with completely unseen classes remains a promising future research direction. 

\begin{table}[t]
\centering
\caption{Evaluation of embodied object detectors on a constrained open-vocabulary object detection task where previously seen class names are replaced with synonyms. All external memory methods are added to the fine-tuned DETIC model, with values inside the parenthesis referring to the change in performance relative to fine-tuned DETIC.}
\begin{tabular}{l|l|l}
\hline
                                            & MP3D                   & Replica                \\ \hline
Implicit Object Memory   & 27.28 (+1.11)   & \textbf{40.93 (+0.90) } \\ \hline
Explicit Object Memory   & \textbf{27.43 (+1.26) }          & 39.92 (-0.09)           \\ \hline
Implicit Pixel Memory    & 25.99 (-0.18)           & 39.26 (-0.77)           \\ \hline
Fine-tuned DETIC & 26.17           & 40.03           \\ \hline
Pre-trained DETIC                            & 13.90                   & 28.34                  \\ \hline
\end{tabular}
\label{table3}
\vspace{-1.5em}
\end{table}

\subsection{Hyper-parameter Selection and Sensitivity}
In all experiments, optimised weighting coefficients $\lambda$ were used to combine external memory features and image features. The sensitivity of the implicit object memory, explicit object memory and implicit pixel memory to these parameters on the Matterport3D test set are shown in Figure \ref{fig:hyperparams}. While there is clear benefit to tuning the weighting coefficients, performance remains strong for each method across a range of values. We also assess the sensitivity of the implicit object memory to the threshold $\tau_{s}$ used to select confident detections for memory update. Performance remains strong across a range of values, showing minor variability when set between 0.2 and 0.5 (Figure \ref{fig:hyperparams}). We persist with a value of 0.3 for all experiments, as this was used to visualise DETIC performance in previous work \cite{detic}.

%% file: Sections/6.0_Robot_Experiments.tex
\section{Robot Experiments}
\subsection{Data Collection}
We deploy our embodied object detector on a mobile robot platform to verify its effectiveness under real-world conditions (Figure \ref{fig:robot}). The platform uses a RealSense D455 camera and an off-the-shelf SLAM system to collect RGB-D and robot pose data at 16Hz. We use HDL Graph SLAM to generate a geometric map of the environment and HDL Localisation to calculate the pose of the robot \cite{slam}. Note that the SLAM system performs geometric mapping only, with all semantic information being generated by the embodied object detector. Three traversals of an office environment (Figure \ref{fig:map}) were collected along a pre-recorded path, two clockwise and one anti-clockwise. The classes \textit{chair, table, sofa, picture, tv monitor, and cushion} were present in the scene, and the locations of some objects were changed between traversals. The embodied object detectors were run locally at approximately 10Hz on a Nvidia GeForce RTX 3090. The data and code for all experiments will be made available after the review process is complete.

\begin{table}[t]
\centering
\caption{Performance of the robot on the object recall task when using different embodied object detectors. Fine-tuned DETIC is used as the base model for the implicit object memory.}
\vspace{-1em}
\begin{tabular}{l|l|l|l}
\hline
                                                 & Precision              & Recall     &  Accuracy      \\ \hline
Implicit Object Memory        & \textbf{0.83}    & 0.61 & \textbf{0.77}        \\ \hline
Fine-tuned DETIC                                 & 0.73 & \textbf{0.63} & 0.74 \\ \hline
Vanilla Training                                 & 0.77 & 0.17 & 0.55 \\ \hline
Pre-trained DETIC                                 & 0.61 & 0.52 & 0.63 \\ \hline
\end{tabular}
\label{table4}
\vspace{-1em}
\end{table}

\subsection{Object Recall Task}
An object recall task is used to evaluate the embodied object detectors without labelling all instances in the collected data. The traversals are split into episodes of 100 images, and the robot is tasked with returning a list of object classes encountered in each episode. An object is classified as ``encountered" by the robot if a detection with class-specific likelihood score greater than 0.3 is returned in 5 consecutive images, with an IoU of 0.6 used to associate detections in consecutive images. If this criteria is not met during the episode, the class is considered ``missing". For each encountered class, the robot must additionally return a single bounding box containing a correctly classified object. In this experiment, the detection with the highest confidence score is simply returned for evaluation.


\subsection{Results}
\begin{figure}
    \begin{minipage}{0.35\columnwidth}
    \centering
    \includegraphics[width=1\columnwidth]{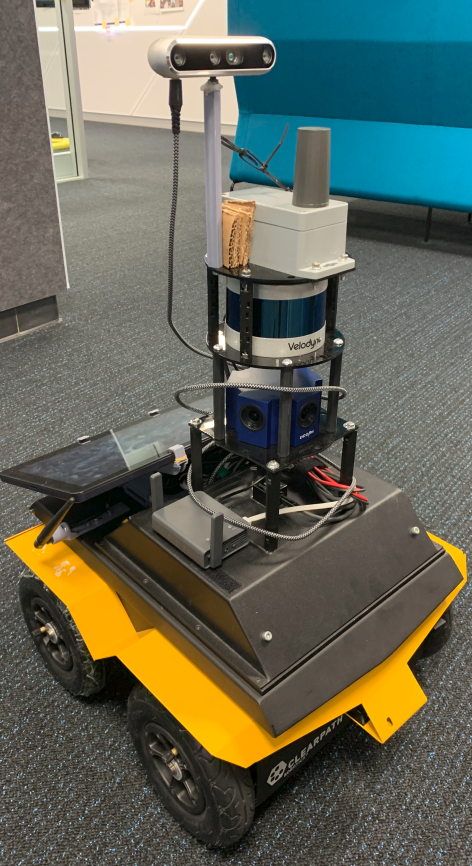}
    \vspace{-1.8em}
    \caption{The mobile robot platform used to evaluate embodied object detection under real-world conditions.
    }
    \label{fig:robot}
    \end{minipage}
    \begin{minipage}{0.05\columnwidth}
        
    \end{minipage}
    \begin{minipage}{0.63\columnwidth}
    \centering
    \vspace{-1em}
    \includegraphics[width=1.0\columnwidth]{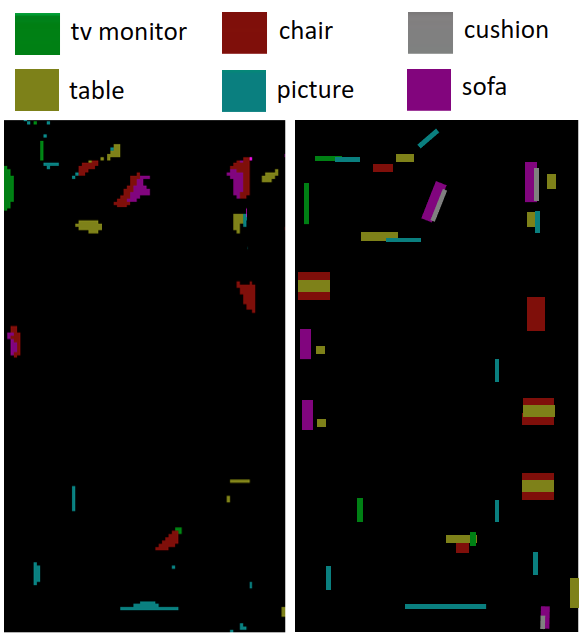}
    \vspace{-2em}
    \caption{Visualisation of the implicit object memory after all three traversals of the scene (left) compared to the ground truth semantic map (right). The implicit object memory is visualised as an explicit semantic map using the approach described in Section IV.D.
    }
    \label{fig:map}
    \end{minipage}
\end{figure}

We report the recall, precision and accuracy of the robot on the object recall task when using different embodied object detectors (Table \ref{table4}). Vanilla training, pre-trained DETIC, fine-tuned DETIC and our implicit object memory are evaluated using models trained in the previous experiments. The implicit object memory maintained throughout the experiment is also visualised as a semantic map in Figure \ref{fig:map}. Note that objects excluded from this semantic map can still be detected by the robot due to the implicit nature of the memory. 

The results show that performing language-image pre-training is valuable for the object recall task, with the fine-tuned DETIC model returning an increase in recall of 0.46 relative to vanilla training (Table \ref{table4}). Furthermore, the addition of implicit object memory to the fine-tuned DETIC model leads to an increase in precision of 0.1 and overall higher accuracy (Table \ref{table4}).  Evidently, the proposed embodied object detector is robust when facing the domain shift, sensor noise and dynamic objects present in real-world deployment. The results also highlight that improved embodied object detection can make a tangible impact on downstream robotic tasks. 

%% file: Sections/7.0_Appendix.tex
\section{Appendix}
\subsection{Class Mappings}
The classes in the Replica dataset differ from those in the Matterport3D dataset used to train the embodied object detectors in this work. To enable testing on the Replica dataset, the following mapping from Replica to Matterport3D classes was used:

\begin{python}
replica_to_lvis_mapping = {
    'chair': 'chair',
    'cushion':'cushion',
    'table':'table',
    'pillow':'cushion',
    'cabinet':'cabinet',
    'shelf':'shelving',
    'rack':'chest_of_drawers',
    'sofa': 'sofa',
    'sink': 'sink',
    'base-cabinet':'cabinet',
    'wall-cabinet':'cabinet',
    'bed':'bed',
    'comforter':'bed',
    'desk': 'table',
    'bathtub': 'bathtub',
    'blinds': 'curtain',
    'curtain': 'curtain',
    'monitor': 'tv_monitor',
    'nightstand': 'table',
    'picture': 'picture',
    'toilet': 'toilet',
    'tv-screen': 'tv_monitor'
    }
\end{python}

To test the embodied object detectors on a constrained version of open-vocabulary detection, ChatGPT was used to provide common synonyms for each class:

\begin{python}
class_synonyms = {
    'bed': 'cot',
    'towel': 'cloth',
    'fireplace': 'hearth',
    'picture': 'painting',
    'cabinet': 'cupboard',
    'toilet': 'lavatory',
    'curtain': 'drapery',
    'table': 'desk',
    'sofa': 'couch',
    'cushion': 'pillow',
    'bathtub': 'tub',
    'chair': 'seat',
    'chest_of_drawers': 'dresser',
    'sink': 'basin',
    'tv_monitor': 'television'
}
\end{python}

%% file: main.bbl
\begin{thebibliography}{66}
\providecommand{\natexlab}[1]{#1}
\providecommand{\url}[1]{\texttt{#1}}
\expandafter\ifx\csname urlstyle\endcsname\relax
  \providecommand{\doi}[1]{doi: #1}\else
  \providecommand{\doi}{doi: \begingroup \urlstyle{rm}\Url}\fi

\bibitem[Anil et~al.(2023)Anil, Dai, Firat, Johnson, Lepikhin, Passos, Shakeri, Taropa, Bailey, Chen, et~al.]{palme}
Rohan Anil, Andrew~M Dai, Orhan Firat, Melvin Johnson, Dmitry Lepikhin, Alexandre Passos, Siamak Shakeri, Emanuel Taropa, Paige Bailey, Zhifeng Chen, et~al.
\newblock Palm 2 technical report.
\newblock \emph{arXiv preprint arXiv:2305.10403}, 2023.

\bibitem[Beeching et~al.(2020)Beeching, Dibangoye, Simonin, and Wolf]{ego_map}
Edward Beeching, Jilles Dibangoye, Olivier Simonin, and Christian Wolf.
\newblock Egomap: Projective mapping and structured egocentric memory for {Deep RL}.
\newblock In \emph{Joint European Conference on Machine Learning and Knowledge Discovery in Databases}, pages 525--540. Springer, 2020.

\bibitem[Campos et~al.(2021)Campos, Elvira, Rodr{\'\i}guez, Montiel, and Tard{\'o}s]{orbslam3}
Carlos Campos, Richard Elvira, Juan J~G{\'o}mez Rodr{\'\i}guez, Jos{\'e}~MM Montiel, and Juan~D Tard{\'o}s.
\newblock Orb-slam3: An accurate open-source library for visual, visual--inertial, and multimap {SLAM}.
\newblock \emph{IEEE Transactions on Robotics}, 37\penalty0 (6):\penalty0 1874--1890, 2021.

\bibitem[Cartillier et~al.(2021)Cartillier, Ren, Jain, Lee, Essa, and Batra]{smnet}
Vincent Cartillier, Zhile Ren, Neha Jain, Stefan Lee, Irfan Essa, and Dhruv Batra.
\newblock Semantic mapnet: Building allocentric semantic maps and representations from egocentric views.
\newblock In \emph{Proceedings of the AAAI Conference on Artificial Intelligence}, volume~35, pages 964--972, 2021.

\bibitem[Chang et~al.(2017)Chang, Dai, Funkhouser, Halber, Niessner, Savva, Song, Zeng, and Zhang]{mp3d}
Angel Chang, Angela Dai, Thomas Funkhouser, Maciej Halber, Matthias Niessner, Manolis Savva, Shuran Song, Andy Zeng, and Yinda Zhang.
\newblock {Matterport3D}: Learning from {RGB-D} data in indoor environments.
\newblock \emph{International Conference on 3D Vision (3DV)}, 2017.

\bibitem[Chaplot et~al.(2021)Chaplot, Dalal, Gupta, Malik, and Salakhutdinov]{seal}
Devendra~Singh Chaplot, Murtaza Dalal, Saurabh Gupta, Jitendra Malik, and Russ~R Salakhutdinov.
\newblock {SEAL}: Self-supervised embodied active learning using exploration and {3d} consistency.
\newblock \emph{Advances in neural information processing systems}, 34:\penalty0 13086--13098, 2021.

\bibitem[Chapman et~al.(2023)Chapman, Dayoub, Browne, and Lehnert]{udaod1}
Nicolas~Harvey Chapman, Feras Dayoub, Will Browne, and Christopher Lehnert.
\newblock Predicting class distribution shift for reliable domain adaptive object detection.
\newblock \emph{IEEE Robotics and Automation Letters}, 8\penalty0 (8):\penalty0 5084--5091, 2023.
\newblock \doi{10.1109/LRA.2023.3290420}.

\bibitem[Chen et~al.(2023{\natexlab{a}})Chen, Zhang, Yang, Peng, and Stiefelhagen]{trans4map}
Chang Chen, Jiaming Zhang, Kailun Yang, Kunyu Peng, and Rainer Stiefelhagen.
\newblock Trans4map: Revisiting holistic bird's-eye-view mapping from egocentric images to allocentric semantics with vision transformers.
\newblock In \emph{Proceedings of the IEEE/CVF Winter Conference on Applications of Computer Vision}, pages 4013--4022, 2023{\natexlab{a}}.

\bibitem[Chen et~al.(2023{\natexlab{b}})Chen, Chabal, Laptev, and Schmid]{object_goal}
Shizhe Chen, Thomas Chabal, Ivan Laptev, and Cordelia Schmid.
\newblock Object goal navigation with recursive implicit maps.
\newblock In \emph{The 2023 IEEE/RSJ International Conference on Intelligent Robots and Systems (IROS 2023)}, 2023{\natexlab{b}}.

\bibitem[Chen et~al.(2017)Chen, Ma, Wan, Li, and Xia]{3d_object_detection}
Xiaozhi Chen, Huimin Ma, Ji~Wan, Bo~Li, and Tian Xia.
\newblock Multi-view {3d} object detection network for autonomous driving.
\newblock In \emph{Proceedings of the IEEE conference on Computer Vision and Pattern Recognition}, pages 1907--1915, 2017.

\bibitem[Chen et~al.(2020)Chen, Cao, Hu, and Wang]{mega}
Yihong Chen, Yue Cao, Han Hu, and Liwei Wang.
\newblock Memory enhanced global-local aggregation for video object detection.
\newblock In \emph{Proceedings of the IEEE/CVF conference on computer vision and pattern recognition}, pages 10337--10346, 2020.

\bibitem[Chen et~al.(2018)Chen, Li, Sakaridis, Dai, and Van~Gool]{udaod2}
Yuhua Chen, Wen Li, Christos Sakaridis, Dengxin Dai, and Luc Van~Gool.
\newblock Domain adaptive faster {r-cnn} for object detection in the wild.
\newblock In \emph{Proceedings of the IEEE Conference on Computer Vision and Pattern Recognition}, pages 3339--3348, 2018.

\bibitem[Deng et~al.(2019{\natexlab{a}})Deng, Hua, Song, Zhang, Xue, Ma, Robertson, and Guan]{ogem}
Hanming Deng, Yang Hua, Tao Song, Zongpu Zhang, Zhengui Xue, Ruhui Ma, Neil Robertson, and Haibing Guan.
\newblock Object guided external memory network for video object detection.
\newblock In \emph{Proceedings of the IEEE/CVF International Conference on Computer Vision}, pages 6678--6687, 2019{\natexlab{a}}.

\bibitem[Deng et~al.(2019{\natexlab{b}})Deng, Pan, Yao, Zhou, Li, and Mei]{relation_networks}
Jiajun Deng, Yingwei Pan, Ting Yao, Wengang Zhou, Houqiang Li, and Tao Mei.
\newblock Relation distillation networks for video object detection.
\newblock In \emph{Proceedings of the IEEE/CVF International Conference on Computer Vision}, pages 7023--7032, 2019{\natexlab{b}}.

\bibitem[Dosovitskiy et~al.(2015)Dosovitskiy, Fischer, Ilg, Hausser, Hazirbas, Golkov, Van Der~Smagt, Cremers, and Brox]{feature_flow_4}
Alexey Dosovitskiy, Philipp Fischer, Eddy Ilg, Philip Hausser, Caner Hazirbas, Vladimir Golkov, Patrick Van Der~Smagt, Daniel Cremers, and Thomas Brox.
\newblock Flownet: Learning optical flow with convolutional networks.
\newblock In \emph{Proceedings of the IEEE International Conference on Computer Vision}, pages 2758--2766, 2015.

\bibitem[Fang et~al.(2020)Fang, Jain, Sarch, Harley, and Fragkiadaki]{move_to_see}
Zhaoyuan Fang, Ayush Jain, Gabriel Sarch, Adam~W Harley, and Katerina Fragkiadaki.
\newblock Move to see better: Self-improving embodied object detection.
\newblock \emph{arXiv preprint arXiv:2012.00057}, 2020.

\bibitem[Feichtenhofer et~al.(2017)Feichtenhofer, Pinz, and Zisserman]{box_1}
Christoph Feichtenhofer, Axel Pinz, and Andrew Zisserman.
\newblock Detect to track and track to detect.
\newblock In \emph{Proceedings of the IEEE International Conference on Computer Vision}, pages 3038--3046, 2017.

\bibitem[Fujitake and Sugimoto(2022)]{transformer_mem}
Masato Fujitake and Akihiro Sugimoto.
\newblock Video sparse transformer with attention-guided memory for video object detection.
\newblock \emph{IEEE Access}, 10:\penalty0 65886--65900, 2022.
\newblock \doi{10.1109/ACCESS.2022.3184031}.

\bibitem[Grinvald et~al.(2019)Grinvald, Furrer, Novkovic, Chung, Cadena, Siegwart, and Nieto]{voxblox}
Margarita Grinvald, Fadri Furrer, Tonci Novkovic, Jen~Jen Chung, Cesar Cadena, Roland Siegwart, and Juan Nieto.
\newblock Volumetric instance-aware semantic mapping and {3D} object discovery.
\newblock \emph{IEEE Robotics and Automation Letters}, 4\penalty0 (3):\penalty0 3037--3044, 2019.

\bibitem[Gu et~al.(2021)Gu, Lin, Kuo, and Cui]{vl_distil}
Xiuye Gu, Tsung-Yi Lin, Weicheng Kuo, and Yin Cui.
\newblock Open-vocabulary object detection via vision and language knowledge distillation.
\newblock In \emph{International Conference on Learning Representations}, 2021.

\bibitem[Guo et~al.(2019)Guo, Fan, Gu, Zhang, Xiang, Prinet, and Pan]{feature_att_2}
Chaoxu Guo, Bin Fan, Jie Gu, Qian Zhang, Shiming Xiang, Veronique Prinet, and Chunhong Pan.
\newblock Progressive sparse local attention for video object detection.
\newblock In \emph{Proceedings of the IEEE/CVF International Conference on Computer Vision}, pages 3909--3918, 2019.

\bibitem[Gupta et~al.(2019)Gupta, Dollar, and Girshick]{lvis}
Agrim Gupta, Piotr Dollar, and Ross Girshick.
\newblock {LVIS}: A dataset for large vocabulary instance segmentation.
\newblock In \emph{Proceedings of the IEEE/CVF conference on computer vision and pattern recognition}, pages 5356--5364, 2019.

\bibitem[Han et~al.(2023)Han, Sun, Ge, Yang, Dong, Zhou, Mao, Peng, and Zhang]{recurrent_fusion}
Chunrui Han, Jianjian Sun, Zheng Ge, Jinrong Yang, Runpei Dong, Hongyu Zhou, Weixin Mao, Yuang Peng, and Xiangyu Zhang.
\newblock Exploring recurrent long-term temporal fusion for multi-view {3d} perception.
\newblock \emph{arXiv preprint arXiv:2303.05970}, 2023.

\bibitem[Han et~al.(2016)Han, Khorrami, Paine, Ramachandran, Babaeizadeh, Shi, Li, Yan, and Huang]{box_4}
Wei Han, Pooya Khorrami, Tom~Le Paine, Prajit Ramachandran, Mohammad Babaeizadeh, Honghui Shi, Jianan Li, Shuicheng Yan, and Thomas~S Huang.
\newblock Seq-{nms} for video object detection.
\newblock \emph{arXiv preprint arXiv:1602.08465}, 2016.

\bibitem[He et~al.(2016)He, Zhang, Ren, and Sun]{resnet}
Kaiming He, Xiangyu Zhang, Shaoqing Ren, and Jian Sun.
\newblock Deep residual learning for image recognition.
\newblock In \emph{Proceedings of the IEEE conference on computer vision and pattern recognition}, pages 770--778, 2016.

\bibitem[Huang et~al.(2023)Huang, Mees, Zeng, and Burgard]{vlmap}
Chenguang Huang, Oier Mees, Andy Zeng, and Wolfram Burgard.
\newblock Visual language maps for robot navigation.
\newblock In \emph{2023 IEEE International Conference on Robotics and Automation (ICRA)}, pages 10608--10615. IEEE, 2023.

\bibitem[Huang and Huang(2022)]{bevdet}
Junjie Huang and Guan Huang.
\newblock Bevdet4d: Exploit temporal cues in multi-camera {3d} object detection.
\newblock \emph{arXiv preprint arXiv:2203.17054}, 2022.

\bibitem[Jiang et~al.(2020)Jiang, Liu, Yang, Liu, Gao, Zhang, Xiang, and Pan]{feature_att_3}
Zhengkai Jiang, Yu~Liu, Ceyuan Yang, Jihao Liu, Peng Gao, Qian Zhang, Shiming Xiang, and Chunhong Pan.
\newblock Learning where to focus for efficient video object detection.
\newblock In \emph{Computer Vision--ECCV 2020: 16th European Conference, Glasgow, UK, August 23--28, 2020, Proceedings, Part XVI 16}, pages 18--34. Springer, 2020.

\bibitem[Kamath et~al.(2021)Kamath, Singh, LeCun, Synnaeve, Misra, and Carion]{mdetr}
Aishwarya Kamath, Mannat Singh, Yann LeCun, Gabriel Synnaeve, Ishan Misra, and Nicolas Carion.
\newblock Mdetr-modulated detection for end-to-end multi-modal understanding.
\newblock In \emph{Proceedings of the IEEE/CVF International Conference on Computer Vision}, pages 1780--1790, 2021.

\bibitem[Kang et~al.(2017{\natexlab{a}})Kang, Li, Xiao, Ouyang, Yan, Liu, and Wang]{box_2}
Kai Kang, Hongsheng Li, Tong Xiao, Wanli Ouyang, Junjie Yan, Xihui Liu, and Xiaogang Wang.
\newblock Object detection in videos with tubelet proposal networks.
\newblock In \emph{Proceedings of the IEEE conference on computer vision and pattern recognition}, pages 727--735, 2017{\natexlab{a}}.

\bibitem[Kang et~al.(2017{\natexlab{b}})Kang, Li, Yan, Zeng, Yang, Xiao, Zhang, Wang, Wang, Wang, et~al.]{box_3}
Kai Kang, Hongsheng Li, Junjie Yan, Xingyu Zeng, Bin Yang, Tong Xiao, Cong Zhang, Zhe Wang, Ruohui Wang, Xiaogang Wang, et~al.
\newblock T-{cnn}: Tubelets with convolutional neural networks for object detection from videos.
\newblock \emph{IEEE Transactions on Circuits and Systems for Video Technology}, 28\penalty0 (10):\penalty0 2896--2907, 2017{\natexlab{b}}.

\bibitem[Kim et~al.(2021)Kim, Koh, Lee, Yang, and Choi]{feature_att_1}
Jaekyum Kim, Junho Koh, Byeongwon Lee, Seungji Yang, and Jun~Won Choi.
\newblock Video object detection using object's motion context and spatio-temporal feature aggregation.
\newblock In \emph{2020 25th International Conference on Pattern Recognition (ICPR)}, pages 1604--1610. IEEE, 2021.

\bibitem[Kirillov et~al.(2023)Kirillov, Mintun, Ravi, Mao, Rolland, Gustafson, Xiao, Whitehead, Berg, Lo, et~al.]{segment_anything}
Alexander Kirillov, Eric Mintun, Nikhila Ravi, Hanzi Mao, Chloe Rolland, Laura Gustafson, Tete Xiao, Spencer Whitehead, Alexander~C Berg, Wan-Yen Lo, et~al.
\newblock Segment anything.
\newblock \emph{arXiv preprint arXiv:2304.02643}, 2023.

\bibitem[Koide et~al.(2019)Koide, Miura, and Menegatti]{slam}
Kenji Koide, Jun Miura, and Emanuele Menegatti.
\newblock A portable three-dimensional lidar-based system for long-term and wide-area people behavior measurement.
\newblock \emph{International Journal of Advanced Robotic Systems}, 16\penalty0 (2):\penalty0 1729881419841532, 2019.

\bibitem[Kotar and Mottaghi(2022)]{interactron}
Klemen Kotar and Roozbeh Mottaghi.
\newblock Interactron: Embodied adaptive object detection.
\newblock In \emph{Proceedings of the IEEE/CVF Conference on Computer Vision and Pattern Recognition}, pages 14860--14869, 2022.

\bibitem[Li et~al.(2022)Li, Wang, Li, Xie, Sima, Lu, Qiao, and Dai]{bevformer}
Zhiqi Li, Wenhai Wang, Hongyang Li, Enze Xie, Chonghao Sima, Tong Lu, Yu~Qiao, and Jifeng Dai.
\newblock Bevformer: Learning bird’s-eye-view representation from multi-camera images via spatiotemporal transformers.
\newblock In \emph{European conference on computer vision}, pages 1--18. Springer, 2022.

\bibitem[Lin et~al.(2020)Lin, Chen, Zhang, Liang, Li, Shan, and Wang]{feature_att_4}
Lijian Lin, Haosheng Chen, Honglun Zhang, Jun Liang, Yu~Li, Ying Shan, and Hanzi Wang.
\newblock Dual semantic fusion network for video object detection.
\newblock In \emph{Proceedings of the 28th ACM international conference on multimedia}, pages 1855--1863, 2020.

\bibitem[Lin et~al.(2014)Lin, Maire, Belongie, Hays, Perona, Ramanan, Doll{\'a}r, and Zitnick]{coco}
Tsung-Yi Lin, Michael Maire, Serge Belongie, James Hays, Pietro Perona, Deva Ramanan, Piotr Doll{\'a}r, and C~Lawrence Zitnick.
\newblock Microsoft coco: Common objects in context.
\newblock In \emph{Computer Vision--ECCV 2014: 13th European Conference, Zurich, Switzerland, September 6-12, 2014, Proceedings, Part V 13}, pages 740--755. Springer, 2014.

\bibitem[Lin et~al.(2017)Lin, Doll{\'a}r, Girshick, He, Hariharan, and Belongie]{fpn}
Tsung-Yi Lin, Piotr Doll{\'a}r, Ross Girshick, Kaiming He, Bharath Hariharan, and Serge Belongie.
\newblock Feature pyramid networks for object detection.
\newblock In \emph{Proceedings of the IEEE conference on computer vision and pattern recognition}, pages 2117--2125, 2017.

\bibitem[Liu et~al.(2021)Liu, Lin, Cao, Hu, Wei, Zhang, Lin, and Guo]{swin}
Ze~Liu, Yutong Lin, Yue Cao, Han Hu, Yixuan Wei, Zheng Zhang, Stephen Lin, and Baining Guo.
\newblock Swin transformer: Hierarchical vision transformer using shifted windows.
\newblock In \emph{Proceedings of the IEEE/CVF International Conference on Computer Vision}, pages 10012--10022, 2021.

\bibitem[Min et~al.(2022)Min, Park, Kim, Park, and Kim]{grounding_with_text}
Seonwoo Min, Nokyung Park, Siwon Kim, Seunghyun Park, and Jinkyu Kim.
\newblock Grounding visual representations with texts for domain generalization.
\newblock In \emph{European Conference on Computer Vision}, pages 37--53. Springer, 2022.

\bibitem[Nilsson et~al.(2021)Nilsson, Pirinen, G{\"a}rtner, and Sminchisescu]{eal_semseg}
David Nilsson, Aleksis Pirinen, Erik G{\"a}rtner, and Cristian Sminchisescu.
\newblock Embodied visual active learning for semantic segmentation.
\newblock In \emph{Proceedings of the AAAI Conference on Artificial Intelligence}, volume~35, pages 2373--2383, 2021.

\bibitem[Patel and Chernova(2022)]{object_modeling}
Maithili Patel and Sonia Chernova.
\newblock Proactive robot assistance via spatio-temporal object modeling.
\newblock In \emph{6th Annual Conference on Robot Learning}, 2022.

\bibitem[Radford et~al.(2021)Radford, Kim, Hallacy, Ramesh, Goh, Agarwal, Sastry, Askell, Mishkin, Clark, et~al.]{clip}
Alec Radford, Jong~Wook Kim, Chris Hallacy, Aditya Ramesh, Gabriel Goh, Sandhini Agarwal, Girish Sastry, Amanda Askell, Pamela Mishkin, Jack Clark, et~al.
\newblock Learning transferable visual models from natural language supervision.
\newblock In \emph{International Conference on Machine Learning}, pages 8748--8763. PMLR, 2021.

\bibitem[Ridnik et~al.(2021)Ridnik, Ben-Baruch, Noy, and Zelnik-Manor]{imagenet21k}
Tal Ridnik, Emanuel Ben-Baruch, Asaf Noy, and Lihi Zelnik-Manor.
\newblock Imagenet-21k pretraining for the masses.
\newblock \emph{arXiv preprint arXiv:2104.10972}, 2021.

\bibitem[Russakovsky et~al.(2015)Russakovsky, Deng, Su, Krause, Satheesh, Ma, Huang, Karpathy, Khosla, Bernstein, et~al.]{imagenetvid}
Olga Russakovsky, Jia Deng, Hao Su, Jonathan Krause, Sanjeev Satheesh, Sean Ma, Zhiheng Huang, Andrej Karpathy, Aditya Khosla, Michael Bernstein, et~al.
\newblock Imagenet large scale visual recognition challenge.
\newblock \emph{International journal of computer vision}, 115:\penalty0 211--252, 2015.

\bibitem[Savva et~al.(2019)Savva, Kadian, Maksymets, Zhao, Wijmans, Jain, Straub, Liu, Koltun, Malik, Parikh, and Batra]{habitat}
Manolis Savva, Abhishek Kadian, Oleksandr Maksymets, Yili Zhao, Erik Wijmans, Bhavana Jain, Julian Straub, Jia Liu, Vladlen Koltun, Jitendra Malik, Devi Parikh, and Dhruv Batra.
\newblock Habitat: A platform for embodied {AI} research.
\newblock In \emph{2019 IEEE/CVF International Conference on Computer Vision (ICCV)}, pages 9338--9346, 2019.
\newblock \doi{10.1109/ICCV.2019.00943}.

\bibitem[Scarpellini et~al.(2023)Scarpellini, Rosa, Morerio, Natale, and Del~Bue]{self_improving}
Gianluca Scarpellini, Stefano Rosa, Pietro Morerio, Lorenzo Natale, and Alessio Del~Bue.
\newblock Self-improving object detection via disagreement reconciliation.
\newblock \emph{arXiv preprint arXiv:2302.10624}, 2023.

\bibitem[Straub et~al.(2019)Straub, Whelan, Ma, Chen, Wijmans, Green, Engel, Mur-Artal, Ren, Verma, Clarkson, Yan, Budge, Yan, Pan, Yon, Zou, Leon, Carter, Briales, Gillingham, Mueggler, Pesqueira, Savva, Batra, Strasdat, Nardi, Goesele, Lovegrove, and Newcombe]{replica}
Julian Straub, Thomas Whelan, Lingni Ma, Yufan Chen, Erik Wijmans, Simon Green, Jakob~J. Engel, Raul Mur-Artal, Carl Ren, Shobhit Verma, Anton Clarkson, Mingfei Yan, Brian Budge, Yajie Yan, Xiaqing Pan, June Yon, Yuyang Zou, Kimberly Leon, Nigel Carter, Jesus Briales, Tyler Gillingham, Elias Mueggler, Luis Pesqueira, Manolis Savva, Dhruv Batra, Hauke~M. Strasdat, Renzo~De Nardi, Michael Goesele, Steven Lovegrove, and Richard Newcombe.
\newblock The {R}eplica dataset: A digital replica of indoor spaces.
\newblock \emph{arXiv preprint arXiv:1906.05797}, 2019.

\bibitem[Sun et~al.(2021)Sun, Hua, Hu, and Robertson]{mamba}
Guanxiong Sun, Yang Hua, Guosheng Hu, and Neil Robertson.
\newblock Mamba: Multi-level aggregation via memory bank for video object detection.
\newblock In \emph{Proceedings of the AAAI Conference on Artificial Intelligence}, volume~35, pages 2620--2627, 2021.

\bibitem[Vaswani et~al.(2017)Vaswani, Shazeer, Parmar, Uszkoreit, Jones, Gomez, Kaiser, and Polosukhin]{attention}
Ashish Vaswani, Noam Shazeer, Niki Parmar, Jakob Uszkoreit, Llion Jones, Aidan~N Gomez, {\L}ukasz Kaiser, and Illia Polosukhin.
\newblock Attention is all you need.
\newblock \emph{Advances in neural information processing systems}, 30, 2017.

\bibitem[Wang et~al.(2018{\natexlab{a}})Wang, Zhou, Yan, and Deng]{feature_flow_2}
Shiyao Wang, Yucong Zhou, Junjie Yan, and Zhidong Deng.
\newblock Fully motion-aware network for video object detection.
\newblock In \emph{Proceedings of the European conference on computer vision (ECCV)}, pages 542--557, 2018{\natexlab{a}}.

\bibitem[Wang et~al.(2018{\natexlab{b}})Wang, Zhou, Yan, and Deng]{feature_flow_5}
Shiyao Wang, Yucong Zhou, Junjie Yan, and Zhidong Deng.
\newblock Fully motion-aware network for video object detection.
\newblock In \emph{Proceedings of the European conference on computer vision (ECCV)}, pages 542--557, 2018{\natexlab{b}}.

\bibitem[Wen et~al.(2020)Wen, Du, Cai, Lei, Chang, Qi, Lim, Yang, and Lyu]{traffic_dataset}
Longyin Wen, Dawei Du, Zhaowei Cai, Zhen Lei, Ming-Ching Chang, Honggang Qi, Jongwoo Lim, Ming-Hsuan Yang, and Siwei Lyu.
\newblock Ua-detrac: A new benchmark and protocol for multi-object detection and tracking.
\newblock \emph{Computer Vision and Image Understanding}, 193:\penalty0 102907, 2020.

\bibitem[Wu et~al.(2019{\natexlab{a}})Wu, Chen, Wang, and Zhang]{SELSA}
Haiping Wu, Yuntao Chen, Naiyan Wang, and Zhaoxiang Zhang.
\newblock Sequence level semantics aggregation for video object detection.
\newblock In \emph{Proceedings of the IEEE/CVF International Conference on Computer Vision}, pages 9217--9225, 2019{\natexlab{a}}.

\bibitem[Wu et~al.(2019{\natexlab{b}})Wu, Kirillov, Massa, Lo, and Girshick]{detectron2}
Yuxin Wu, Alexander Kirillov, Francisco Massa, Wan-Yen Lo, and Ross Girshick.
\newblock Detectron2.
\newblock \url{https://github.com/facebookresearch/detectron2}, 2019{\natexlab{b}}.

\bibitem[Zang et~al.(2022)Zang, Li, Zhou, Huang, and Loy]{open_vocab_detr}
Yuhang Zang, Wei Li, Kaiyang Zhou, Chen Huang, and Chen~Change Loy.
\newblock Open-vocabulary detr with conditional matching.
\newblock In \emph{European Conference on Computer Vision}, pages 106--122. Springer, 2022.

\bibitem[Zang et~al.(2023)Zang, Li, Han, Zhou, and Loy]{contextual_detection}
Yuhang Zang, Wei Li, Jun Han, Kaiyang Zhou, and Chen~Change Loy.
\newblock Contextual object detection with multimodal large language models.
\newblock \emph{arXiv preprint arXiv:2305.18279}, 2023.

\bibitem[Zeng et~al.(2018)Zeng, Zhou, Jenkins, and Desingh]{ctmap}
Zhen Zeng, Yunwen Zhou, Odest~Chadwicke Jenkins, and Karthik Desingh.
\newblock Semantic mapping with simultaneous object detection and localization.
\newblock In \emph{2018 IEEE/RSJ International Conference on Intelligent Robots and Systems (IROS)}, pages 911--918. IEEE, 2018.

\bibitem[Zhao et~al.(2022)Zhao, Zhang, Schulter, Zhao, Vijay~Kumar, Stathopoulos, Chandraker, and Metaxas]{exploiting_unlabeled}
Shiyu Zhao, Zhixing Zhang, Samuel Schulter, Long Zhao, BG~Vijay~Kumar, Anastasis Stathopoulos, Manmohan Chandraker, and Dimitris~N Metaxas.
\newblock Exploiting unlabeled data with vision and language models for object detection.
\newblock In \emph{European Conference on Computer Vision}, pages 159--175. Springer, 2022.

\bibitem[Zhong et~al.(2022)Zhong, Yang, Zhang, Li, Codella, Li, Zhou, Dai, Yuan, Li, et~al.]{regionclip}
Yiwu Zhong, Jianwei Yang, Pengchuan Zhang, Chunyuan Li, Noel Codella, Liunian~Harold Li, Luowei Zhou, Xiyang Dai, Lu~Yuan, Yin Li, et~al.
\newblock Regionclip: Region-based language-image pretraining.
\newblock In \emph{Proceedings of the IEEE/CVF Conference on Computer Vision and Pattern Recognition}, pages 16793--16803, 2022.

\bibitem[Zhou et~al.(2021)Zhou, Koltun, and Kr{\"a}henb{\"u}hl]{centernet2}
Xingyi Zhou, Vladlen Koltun, and Philipp Kr{\"a}henb{\"u}hl.
\newblock Probabilistic two-stage detection.
\newblock \emph{arXiv preprint arXiv:2103.07461}, 2021.

\bibitem[Zhou et~al.(2022)Zhou, Girdhar, Joulin, Kr{\"a}henb{\"u}hl, and Misra]{detic}
Xingyi Zhou, Rohit Girdhar, Armand Joulin, Philipp Kr{\"a}henb{\"u}hl, and Ishan Misra.
\newblock Detecting twenty-thousand classes using image-level supervision.
\newblock In \emph{European Conference on Computer Vision}, pages 350--368. Springer, 2022.

\bibitem[Zhu et~al.(2023)Zhu, Kapoor, Min, Han, Li, Geng, Neubig, Bisk, Kembhavi, and Weihs]{embodied_exploration}
Hao Zhu, Raghav Kapoor, So~Yeon Min, Winson Han, Jiatai Li, Kaiwen Geng, Graham Neubig, Yonatan Bisk, Aniruddha Kembhavi, and Luca Weihs.
\newblock Excalibur: Encouraging and evaluating embodied exploration.
\newblock In \emph{Proceedings of the IEEE/CVF Conference on Computer Vision and Pattern Recognition}, pages 14931--14942, 2023.

\bibitem[Zhu et~al.(2017)Zhu, Wang, Dai, Yuan, and Wei]{feature_flow_1}
Xizhou Zhu, Yujie Wang, Jifeng Dai, Lu~Yuan, and Yichen Wei.
\newblock Flow-guided feature aggregation for video object detection.
\newblock In \emph{Proceedings of the IEEE International Conference on Computer Vision}, pages 408--417, 2017.

\bibitem[Zhu et~al.(2018)Zhu, Dai, Yuan, and Wei]{feature_flow_3}
Xizhou Zhu, Jifeng Dai, Lu~Yuan, and Yichen Wei.
\newblock Towards high performance video object detection.
\newblock In \emph{Proceedings of the IEEE conference on computer vision and pattern recognition}, pages 7210--7218, 2018.

\end{thebibliography}
